%% file: neurips_2023.tex
\documentclass{article}

\usepackage[numbers]{natbib}
\usepackage[final]{neurips_2023}

\usepackage[utf8]{inputenc} 
\usepackage[T1]{fontenc}    
\usepackage{hyperref}       
\usepackage{url}            
\usepackage{booktabs}       
\usepackage{amsfonts}       
\usepackage{nicefrac}       
\usepackage{microtype}      
\usepackage{xcolor}         

\usepackage{graphicx}
\usepackage{caption}
\usepackage{subfig}
\usepackage{adjustbox}
\usepackage{csvsimple}
\usepackage{pgfplots}
\usepackage{multirow}
\usepackage{makecell}
\usetikzlibrary{patterns}

\title{Game Solving with Online Fine-Tuning}

\author{
    Ti-Rong Wu,\textsuperscript{1}\thanks{These authors contributed equally.}\hspace{4pt} Hung Guei,\textsuperscript{1}$^{\ast}$ Ting Han Wei,\textsuperscript{2} Chung-Chin Shih,\textsuperscript{1,3} Jui-Te Chin,\textsuperscript{3} I-Chen Wu\textsuperscript{3,4}\\
    \vspace{1pt}\\
    \textsuperscript{\rm 1}Institute of Information Science, Academia Sinica, Taiwan\\
    \textsuperscript{\rm 2}Department of Computing Science, University of Alberta, Canada\\
    \textsuperscript{\rm 3}Department of Computer Science, National Yang Ming Chiao Tung University, Taiwan\\
    \textsuperscript{\rm 4}Research Center for Information Technology Innovation, Academia Sinica, Taiwan\\
    \vspace{1pt}\\
    \small{\texttt{tirongwu@iis.sinica.edu.tw, hguei@iis.sinica.edu.tw, tinghan@ualberta.ca}} \\
    \small{\texttt{rockmanray.cs02@nycu.edu.tw, pikachin.cs10@nycu.edu.tw, icwu@cs.nctu.edu.tw}}
}

\begin{document}

\maketitle

\input{main}

\bibliography{neurips_2023.bib}
\bibliographystyle{unsrtnat}

\newpage
\input{appendix}

\end{document}

%% file: main.tex
\begin{abstract}
Game solving is a similar, yet more difficult task than mastering a game.
Solving a game typically means to find the game-theoretic value (outcome given optimal play), and optionally a full strategy to follow in order to achieve that outcome.
The AlphaZero algorithm has demonstrated super-human level play, and its powerful policy and value predictions have also served as heuristics in game solving.
However, to solve a game and obtain a full strategy, a winning response must be found for all possible moves by the losing player.
This includes very poor lines of play from the losing side, for which the AlphaZero self-play process will not encounter.
AlphaZero-based heuristics can be highly inaccurate when evaluating these out-of-distribution positions, which occur throughout the entire search.
To address this issue, this paper investigates applying online fine-tuning while searching and proposes two methods to learn tailor-designed heuristics for game solving.
Our experiments show that using online fine-tuning can solve a series of challenging 7x7 Killall-Go problems, using only 23.54\% of computation time compared to the baseline without online fine-tuning.
Results suggest that the savings scale with problem size.
Our method can further be extended to any tree search algorithm for problem solving.
Our code is available at https://rlg.iis.sinica.edu.tw/papers/neurips2023-online-fine-tuning-solver.
\end{abstract}

\section{Introduction}
\label{sec:introduction}

\textit{Playing} and \textit{solving} strategic games have served as drivers and major milestones \cite{van2002games} in artificial intelligence research.
To master such games, the objective is often designed to optimize on the objective of maximizing the probability of winning.
In the past several decades, researchers made significant progress in game playing, reaching super-human playing levels in many domains. 
Successful examples include Chinook (checkers) \cite{schaeffer1997one}, Deep Blue (chess) \cite{campbell2002deep}, AlphaGo (Go) \cite{silver2016mastering}, and AlphaStar (StarCraft II) \cite{vinyals2019grandmaster}. 
Furthermore, AlphaZero \cite{silver2017mastering,silver2018general} and MuZero \cite{schrittwieser2020mastering} even boast generality by mastering a variety of games without requiring expert human knowledge. 
Although these learning-based agents have progressed dramatically in playing strength, there are no guarantees that their decisions are always correct \cite{Gleave2020Adversarial,lan2022are} in terms of game-theoretic value, which is defined as the outcome of the game given optimal play for both players. 
Game solving is this pursuit of finding game-theoretic values.

Game solving is a more difficult challenge than game playing.
Many seemingly simple games have astronomically large state spaces, with no simple way of exploring this space.
Here is where advancements in game playing can aid game solving.
Strong agents are commonly leveraged to evaluate positions, providing guidance and reducing the search space significantly.
For example, the checkers program Chinook claimed to have reached super-human levels as early as 1996 \cite{schaeffer1997one}, then about 10 years later, played an instrumental role in the proof that checkers is a drawn game \cite{schaeffer2007checkers}. 
Similarly, contemporary learning-based approaches such as AlphaZero are widely used to help reduce the search space for game solving. 
Game solvers that utilized AlphaZero include Hex \cite{gao2017focused}, Go \cite{shih2022novel}, Killall-Go \cite{wu2022alphazero}, and the Rubik's cube \cite{mcaleer2018solving, agostinelli2019solving}. 
Such approaches are not limited to applications in games but extend to other non-game fields, like automated theorem proving \cite{lample2022hypertree}. 

However, a major issue still exists when using learning-based approaches to aid game solving. 
In the two-player, zero-sum setting, a simple description for a proof involves verifying that there is a winning move for the winner, for all possibilities played by the losing side; i.e. no matter how the loser plays, the winner must be able to respond correctly.
However, most learning-based agents are trained along a strong line of play by both players, with some exploration to nearby states for robustness.
Using AlphaZero as an example, training samples are generated via self-play by the best version of itself up to that point.
Learning-based methods are powerful in that they generalize for previously unseen positions, but accuracy tends to drop the further you stray from training samples.
To verify all possibilities on the losing side, the vast majority of positions we must evaluate during the search for a proof are therefore out-of-distribution.
To illustrate, AlphaZero-like networks have been shown to make inconsistent or completely incorrect evaluations, simply by adding two meaningless stones to a position \cite{lan2022are}.
In another example, in the attempt to solve one of the hardest Go life-and-death (L\&D) problems from the famous book Igo Hatsuyoron,\footnote{Igo Hatsuyoron is a classic collection of L\&D problems in Go, which demand complex calculations to solve. L\&D problems are puzzles that test your ability to identify the safety of specific pieces in a given game position.} all AlphaZero-like programs failed. It was hypothesized that this was because these highly specific problems are rarely encountered during training \cite{wu2019hardest}.

This paper proposes applying online fine-tuning methods to train learning-based systems while solving games. 
In our proposed methods, during game solving, an online trainer is added so that the learned heuristic is improved as new data that is relevant to the solving task is generated. 
This is done by utilizing new information such as \textit{solved} and \textit{critical} positions in the current solver search tree.
The trainer therefore can learn better heuristics dynamically, that are particularly fine-tuned for the upcoming evaluations.
Experiments are conducted on 16 challenging 7x7 Killall-Go three-move openings, shown in Figure \ref{fig:openings}.
We develop a distributed game solver with online fine-tuning, that is built upon the state-of-the-art 7x7 Killall-Go solver \cite{shih2022novel}.
Experiment results show that the online fine-tuning solver can greatly reduce the search space by a factor of 4.61 on average. 
Namely, it searches only 21.69\% of nodes, using 23.54\% of the computation time, as compared to the offline solver.
Most importantly, for larger problems, the online fine-tuning solver performs significantly faster than that without, which implies that our method scales with problem size.

\section{Background}
\label{sec:background}

\subsection{Game solvers}
\label{sec:bg_solving_games}

A two-player zero-sum game is considered \textit{solved} if we know of a winning strategy for either player which guarantees a winning outcome,\footnote{We only consider ``weak solutions'' \cite{van2002games} in this paper, where different opening positions are treated as independent sub-games. Draws are also not considered, but can be determined via two searches, one for each player. If both outcomes are losses, then it must be a draw.} regardless of how the opponent plays; i.e. the player must have at least one action that leads to a win, for all actions by the opponent. 
A winning strategy is often represented as an AND-OR tree called a \textit{solution tree} \cite{pijls2001game}, where the game positions with the winner to move are represented by OR-nodes, and those for the opponent by AND-nodes.
Leaf nodes in a solution tree are all terminal positions where the outcome is a win.


A \textit{solver} is a program that does a proof search and can identify a winning strategy or a solution tree, if found. 
Solvers often rely on heuristic tree search for games with large and complex game state spaces. 
Search algorithms such as alpha-beta search \cite{stockman1979minimax}, proof number search (PNS) \cite{allis1994proof}, or Monte Carlo tree search (MCTS) \cite{winands2008monte, cazenave2010score} have all been shown to be successful.
In addition, previous research has shown that none of the algorithms dominates the others \cite{ewalds2012playing, wei2015software, wu2022alphazero}.

\subsection{Distributed game solver}
\label{sec:bg_distributed_solving_games}

In cases where search spaces are too large for a single instance solver under reasonable time and memory constraints, multiple solvers are often run in parallel, forming a distributed computing system, to scale up the solving process.
Examples of games solved by distributed computing include checkers \cite{schaeffer2007checkers}, heads-up limit hold'em poker \cite{bowling2015hulhesolved}, breakthrough \cite{saffidine2011solving}, Hex \cite{henderson2009solving}, and Connect6 \cite{wei2015software,wu2012job}.

These distributed game solving/analysis systems, also known as \textit{distributed game solvers}, have been presented commonly with two components, a \textit{manager} and a set of \textit{workers}. 
A manager divides the overall problem into smaller sub-problems, keeping only the initial portion of the search tree -- the beginning of the game -- in memory. 
As this search tree is expanded, the manager may decide to offload analysis of specific positions to its workers.
These offloaded sub-problems are also called \textit{jobs} \cite{wei2015software,wu2012job}. 
A worker computes jobs by taking as input, a specific position and any relevant parameters (e.g. time limits), then outputs either a solved or heuristic value for that position.
A worker can be a single solver, a game engine, or even a combination of both. 

For example, in the checkers proof, Chinook and another depth-first proof-number search \cite{nagai2002df} solver were combined as a heuristic.
From the perspective of the manager, a solved job result, such as a \textit{proven} win, loss, or draw, can be thought of as a terminal node in its solution tree. Unsolved jobs also provide useful information, such as heuristic values to determine \textit{likely win} or \textit{likely loss}, to guide further tree expansions, with the worker acting as a relatively expensive and accurate heuristic.
Similar to the checkers proof, Connect6 openings have also been solved by encapsulating the task of solving and playing a position into a single job, which was then dispatched by a manager to a set of workers \cite{wei2015software,wu2012job,chen2014job}. 
In all examples listed above, a centralized scheme is used where one manager coordinates between dozens to hundreds of workers.

\subsection{Proof Cost Network}
\label{sec:bg_pcn}
When using neural networks as heuristics in solving, recent research points out that there is room for improvement when using the value network learned from the AlphaZero algorithm\cite{agostinelli2019solving, wu2022alphazero}. 
In a search tree, when several actions can reach a winning outcome, AlphaZero-trained networks have no preference for choosing one that wins fastest. This can increase the amount of computation significantly. 

To address this challenge, the Proof Cost Network (PCN) \cite{wu2022alphazero} predicts a proof cost value, rather than a win rate. 
The cost value represents a logarithmically-normalized estimate of the number of nodes that are required to solve the position.
Specifically, PCN adopts the AlphaZero training process and generates self-play games using the cost value to guide the MCTS toward faster winning moves.
These self-play games are then used to update PCN's cost values.
The resulting network will focus the proof search on actions with minimal cost. 
Experiments show that the proof cost value is highly correlated to problem difficulty, and can significantly improve solving capability. 

\section{Game solver with online fine-tuning}
\label{sec:online_method}

This section describes our methods for applying online fine-tuning to game solving.
We chose an MCTS-based solver due to its popularity when integrating AlphaZero networks as heuristics \cite{agostinelli2019solving, shih2022novel, wu2022alphazero}.
However, it is worth noting that the methods presented in this paper are search-independent and can be readily applied to other search algorithms such as alpha-beta search or PNS.

\subsection{Distributed game solver}
\label{sec:method_distributed_solver}


Our distributed game solver consists of a PCN serving as its heuristic, a \textit{manager}, and a set of \textit{workers}.
The manager maintains an MCTS rooted at the position to be solved. 
During the proof search, the manager follows PUCT \cite{rosin2011multi} selection to traverse from the root to a leaf node.
Next, the PCN estimates the cost of the selected leaf node, denoted by $v_l$. $v_l$ is a heuristic value representing the log estimated number of positions that must be examined to solve this node.
If the value is larger than a designated threshold, i.e. $v_l \geq v_{thr}$, its proof cost is considered too high to warrant a job.
The manager will then continue to follow MCTS, expanding the node and backpropagating $v_l$ to the root.
Alternatively, if $v_l<v_{thr}$, the leaf node is highly likely to be solved outright by a worker, at which point a job is created. 
Job granularity is therefore controlled with $v_{thr}$. 
Larger $v_{thr}$ generates more difficult jobs with higher failure rates, while smaller $v_{thr}$ leads to easier but more numerous jobs.
A balanced $v_{thr}$ should be set according to the game instance and worker capabilities. 

\begin{figure}[t]
    \begin{minipage}[b]{0.44\linewidth}
        \centering
        \includegraphics[width=\linewidth]{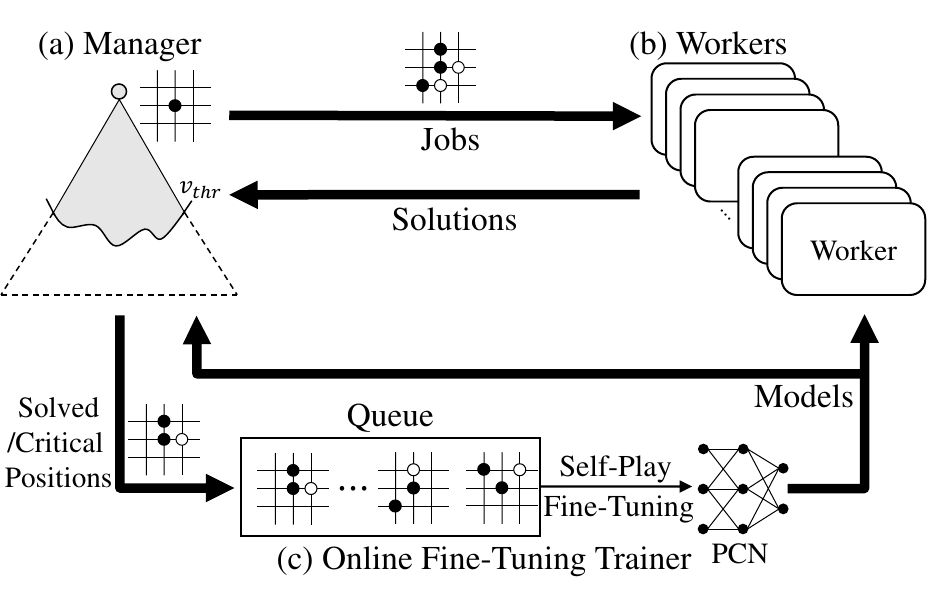}
        \captionsetup[subfigure]{justification=centering}
        \caption{The online fine-tuning game solver architecture.}
        \label{fig:architecture}
    \end{minipage}
    \hspace*{0.1em}
    \begin{minipage}[b]{0.54\linewidth}
        \centering
        \resizebox{0.7\linewidth}{!}{\includegraphics[width=\linewidth]{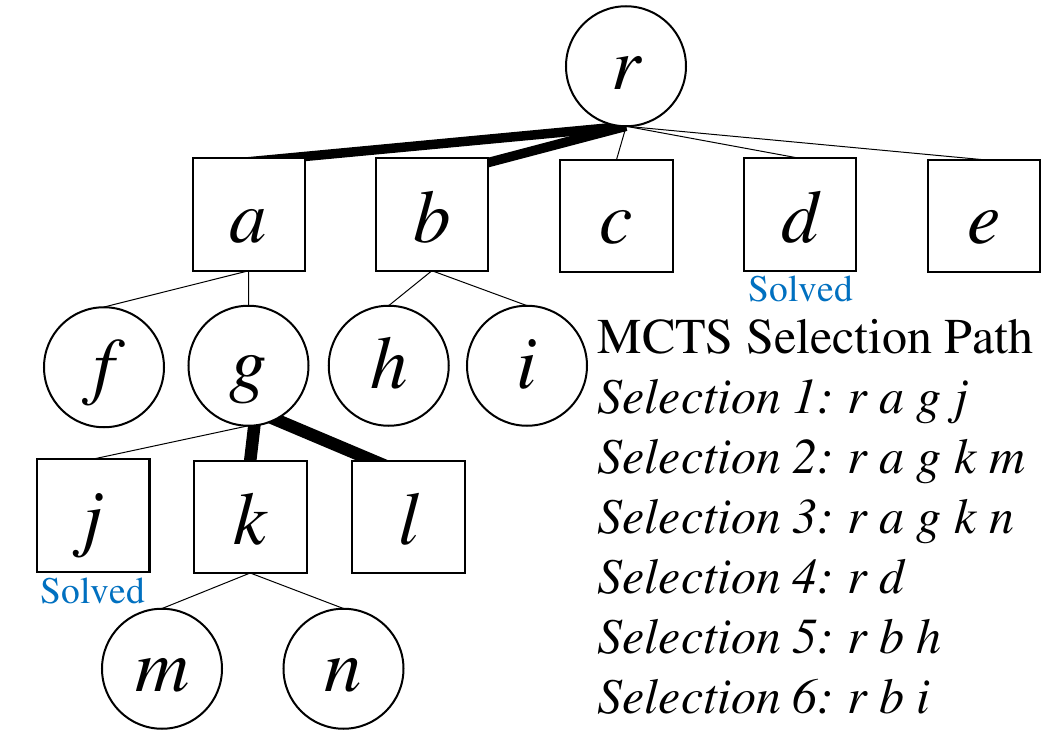}}
        \captionsetup[subfigure]{justification=centering}
        \caption{A manager AND-OR tree with six MCTS selection paths. Positions $d$ and $j$ are solved in this tree.}
        \label{fig:manager-tree-illustration}
    \end{minipage}
    \label{fig:arch_and_manager_merge}
\end{figure}

Workers are game solving programs that are limited by specific constraints, say, a given time limit.
To keep the heuristic consistent during the proof search, workers use the same PCN weights as the manager. 
If the job is solved within the given constraint, the worker returns the result, either a win or a loss, back to the manager; otherwise, it returns an unknown. 
Once the manager receives the job result, it updates the search tree accordingly. 
For unsolved jobs, the manager expands the nodes that generated the corresponding jobs. 
The interaction between the manager and the workers is shown between Figure \ref{fig:architecture}(a) and \ref{fig:architecture}(b).

\subsection{Online Fine-Tuning Trainer}
\label{sec:method_online_trainer}

The Online Fine-Tuning Trainer (OFT) maintains the PCN during the proof search so that the manager and workers have access to ever improving heuristics. 
Without online fine-tuning, both the manager and workers simply use a fixed PCN, denoted by $\theta_0$, trained via the AlphaZero self-play process. 
The OFT starts with $\theta_0$, then fine-tunes the weights via updates during the proof search. 
To do this, the manager picks out \textit{solved} and/or \textit{critical} positions in its search tree, adds them to the list of training samples, then the OFT uses them to perform self-play and training. The fine-tuned PCN ($\theta_1, \theta_2, ..., \theta_t, ...$) can then be used to further guide the manager and workers towards a faster proof. 
The manager and workers update to the most recent $\theta_t$ immediately when a new PCN checkpoint is trained by the OFT.
The above iterative process is shown in Figure \ref{fig:architecture}.
Details are provided in the following sections.

\subsubsection{Online fine-tuning trainer with solved positions}
\label{sec:method_online_trainer_sp}
During the proof search, many previously unsolved positions may become \textit{solved} in the manager's search tree. 
This new information can be used by the OFT to improve the accuracy of the PCN.
Figure \ref{fig:manager-tree-illustration} provides an example of a manager's AND-OR search tree and six recent selection paths.
In this example, positions $j$ and $d$ are marked as solved and sent to the OFT after the first and fourth selection, respectively.
The OFT maintains a queue that stores these solved positions, as shown in Figure \ref{fig:architecture}(c).
Self-play games are generated as in normal PCN training \cite{wu2022alphazero}.
However, in the optimization phase, the OFT randomly samples training data not only from the generated self-play games, but also from the queue of solved positions.
For these solved positions, the cost values are always set to zero (i.e. solved to be a win, from the perspective of the OR-player), since no nodes need to be examined to solve the position.
The OFT only samples 10\% of training data from the solved queue during optimization to avoid overfitting, where the remaining 90\% are sampled from self-play games.
In addition, the queue only stores the most recent 1,000 solved positions received from the manager.
During self-play, when using $\theta_t$ to evaluate positions that are solved by the manager, it is highly likely to predict costs close to zero.
From the AND-player's perspective, it favors moves that lead to larger costs to delay the OR-player's victory.
Therefore, self-play naturally explores positions which have not yet been solved in the manager's search tree.

\subsubsection{Online fine-tuning trainer with critical positions}
\label{sec:method_online_trainer_cp}
Other than solved positions, we can also improve the PCN with specific positions of interest chosen from the manager's current search tree.
Positions are considered \textit{critical} if they are selected in the most recent MCTS iterations in the manager.
For example, in the first selection in Figure \ref{fig:manager-tree-illustration}, all positions $r$, $a$, $g$, and $j$ in the MCTS selection path are considered critical positions.
During self-play, the trainer randomly chooses one critical position, and performs self-play starting from that position.
Thus, $\theta_t$ can provide more accurate predictions for positions that proof search is currently exploring.

A more selective process can be used to improve the quality of critical positions.
First, we can omit $r$, since self-play from $r$ is already performed to train $\theta_0$.
Ideally, we would prefer to focus on deeper unsolved positions.
To achieve this, we only consider the leaf position in the selection path as critical.
Also, the OFT maintains a queue in which only the recent 1,000 critical positions are stored.
This way, the OFT can focus on the most urgent positions which are likely to be solved soon.
As these positions are also usually sent to the workers (if the PCN value $v \leq v_{thr}$), the workers can also take advantage of $\theta_t$.
Next, we can omit leaf positions solved solely by the manager; i.e. leaf nodes that were solved not as jobs.
For example, $j$ will not be considered critical in the first selection in Figure \ref{fig:manager-tree-illustration}.
Since $j$ is already solved, it is not necessary to perform self-play from that position.
Only $m$, $n$, $h$, and $i$ will be sent to the OFT as critical positions in the second, third, fifth, and sixth selection, respectively.
Note that $r$, $a$, and $g$ were critical positions before $b$ and $k$ became critical, since the parent nodes are always expanded before their children.
Thus, the trainer will gradually fine-tune the PCN by focusing only on deeper critical positions to help avoid redundancy during fine-tuning.


In summary, the pre-trained $\theta_0$ learns general heuristics by exploring from empty games, while the online $\theta_t$ refines its heuristics for specific positions of interest. 
As a side note, the fine-tuning process is related to the catastrophic forgetting phenomenon \cite{mccloskey1989catastrophic}, as the focus is shifted from one part of the proof search to another. 
Interestingly, forgetting is not only acceptable in this context, but probably even preferred, because the heuristic only needs to be accurate for the part of the search space the manager is currently working on.
Additionally, the two proposed methods are independent and can be combined. 
We evaluate these methods in our experiments.

\subsection{Manager job assignment improvements}
\label{sec:method_job_assignment}

Job assignment refers to the manager's responsibility of dividing the overall problem into distinct jobs.
Better job assignment schemes can eliminate redundancy and improve parallelism.
We propose three techniques to further improve the efficiency of job assignment, which we call \textit{virtual solving}, \textit{top-$k$ selection}, and \textit{AND-player job assignment}.

\textbf{Virtual solving.} When a job is assigned to workers, we assume that the job result will be solved, even before it is actually returned by a worker.
The \textit{virtually solved} outcome is backpropagated as a normal job outcome. 
This technique has similar concepts to the \textit{virtual loss} \cite{chaslot2008parallel}, \textit{virtual win} \cite{wu2012job}, and \textit{Young Brothers Wait Concept} (YBWC) \cite{feldmann1989ybwc}, which were used to avoid repeatedly searching superfluous nodes during the proof search.
For example, in Figure \ref{fig:manager-tree-illustration}, assume the manager selects a path from $r$ to a leaf $h$ and assigns the job to a worker, at which point $h$ is immediately marked as virtually solved.
Its parent node $b$, an OR-node, is then also marked as virtually solved.
Furthermore, if nodes $a$, $c$, and $e$ ($d$ is solved already) are all solved or virtually solved, their parent $r$ will also be marked as virtually solved.
When the job result returns, the manager reverts the virtually solved markers and updates the status of all nodes accordingly.
The virtual solving technique can provide a highly efficient job assignment scheme, in that the manager search tree can be nearly the same as the solution tree if most virtually solved nodes are indeed winning. 

\textbf{Top-$k$ selection.} We exploit the fact that all child nodes must be solved for every AND-node to improve parallelism.
At each AND-node, we select uniformly at random among the top $k$ unsolved children that are likely to be sent off as jobs eventually, i.e. those with the top $k$ highest PUCT scores.
For example, in Figure \ref{fig:manager-tree-illustration}, assume $k=2$ and nodes $a$ and $b$ are the top two children of AND-node $r$; nodes $k$ and $l$ are the top two for $g$. Note that we omit node $j$ because it is already solved.
At node $r$, the manager selects between $a$ and $b$ with equal probability. 
Note that selections at OR-nodes remain unchanged. 
In addition, we only apply top-$k$ selection when the simulation count of the AND-node is larger than $k$.
Top-$k$ selection improves parallelism by allowing the manager to assign more jobs simultaneously, when it is combined with virtual solving.
We use $k=4$ in our experiments.

\textbf{AND-player job assignment.} We only distribute AND-nodes as jobs, i.e. OR-nodes are never assigned and are directly expanded in the manager.
For example, in Figure \ref{fig:manager-tree-illustration}, the OR-node $b$ is not assigned as a job even if $v_b < v_{thr}$. 
The manager creates the AND-node $h$ from the leaf node $b$, then assigns it to a worker as a job. 
The underlying intuition is that assuming the PCN policy head output is accurate as a move ordering heuristic, the first guess will often be the move that leads to a solution for OR-nodes.
Therefore, by skipping OR-nodes job assignment entirely, the manager gains a 1-ply look ahead.
In practice, all three job assignment schemes are applied simultaneously.

\section{Experiments}
\label{sec:experiments}

We demonstrate our online fine-tuning game solver by solving several three-move 7x7 Killall-Go openings. 
7x7 Killall-Go is a variant of Go, where the rules are the same except that: (a) Black places two stones initially, and (b) Black wins if all white stones are killed; otherwise, White wins. 
Since White aims to live, winning specific openings for this variant is equivalent to solving a L\&D problem. 
Many Go experts believe that 7x7 Killall-Go is a win for White. 
So far, no proof has been published yet.
In this paper, we only focus on weakly solved games \cite{Allis1994SearchingFS} in which White wins. 
Thus, White is considered the OR-player throughout.

\subsection{The 7x7 Killall-Go solver}
\label{sec:exp_7x7_killall}

We build our 7x7 Killall-Go solver upon an AlphaZero training framework \cite{wu2023minizero}.
First, we pre-train a PCN $\theta_0$ \cite{wu2022alphazero} to serve as heuristics for the game solver (starting from an empty board). 
We incorporate the Gumbel AlphaZero algorithm \cite{danihelka2021policy} into PCN training, since it performs equivalently well even with a small simulation count.
This reduces the computation cost for online fine-tuning without compromising accuracy. 
The pre-training took around 52 1080Ti GPU-hours. 
Next, we incorporate several useful techniques into the solver to accelerate solving.
This includes relevance zone-based search (RZS) \cite{shih2022novel}, zone pattern tables \cite{shih2023local}, and GHI handling to deal with cycles in Go \cite{kishimoto2004general}.
This solver is then used as workers in a distributed game solver.
The manager is also based on the above solver, with the job assignment techniques added, as described in subsection \ref{sec:method_job_assignment}. The OFT is similar to the PCN pre-training, but with fine-tuning as described in subsection \ref{sec:method_online_trainer}.

Two kinds of distributed game solvers are considered for our experiments.
The \textit{baseline solver} uses the manager and worker only, while using a pre-trained, fixed $\theta_0$ as the heuristic throughout the whole proof search. 
In contrast, the \textit{online fine-tuning solver} uses the OFT to fine-tune the PCN heuristic dynamically during the proof search. 
In addition, we consider three variations of online fine-tuning solvers using solved positions (SP), critical positions (CP), and a combination of both (SP+CP).
Both solvers use $v_{thr}=16.5$ for the manager job granularity.\footnote{We choose $v_{thr}=16.5$ according to the experiments on different PCN thresholds, as shown in the appendix.}
For fairness, we ran both solvers on 9 1080Ti GPUs.
The baseline solver uses one GPU for the manager and eight GPUs shared among workers. 
For the online fine-tuning solver, the manager and trainer each uses one GPU, while workers share the remaining seven GPUs.
Detailed implementations and other machine configuration details are specified in the appendix.

\begin{figure}[h]
    \captionsetup[subfigure]{justification=centering}
    \centering
    \subfloat[Jump (J)]{
        \includegraphics[width=0.15\columnwidth]{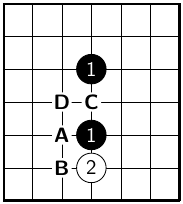}
        \label{fig:openings_jump}
    }
    \hspace*{1em}
    \subfloat[Knight's move (K)]{
        \includegraphics[width=0.15\columnwidth]{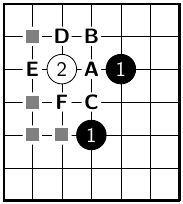}
        \label{fig:openings_knight}
    }
    \hspace*{1em}
    \subfloat[Diagonal jump (D)]{
        \includegraphics[width=0.15\columnwidth]{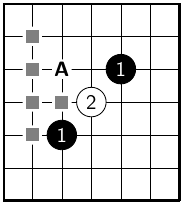}
        \label{fig:openings_diagonal_jump}
    }
    \hspace*{1em}
    \subfloat[Stretch (S)]{
        \includegraphics[width=0.15\columnwidth]{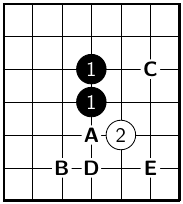}
        \label{fig:openings_stretch}
    }
    \caption{Four 7x7 Killall-Go opening groups, including (a) four openings, JA-JD; (b) six openings, KA-KF; (c) one opening, DA; (d) five openings, SA-SE.}
    \label{fig:openings}
\end{figure}

We select a set of three-move openings based on recommendations from experts, including a professional 9-dan player. 
These openings can be classified into four groups, named after their commonly shared first move opening: \textit{jump} (J), \textit{knight's move} (K), \textit{diagonal jump} (D), and \textit{stretch} (S), shown in Figure \ref{fig:openings_jump}, \ref{fig:openings_knight}, \ref{fig:openings_diagonal_jump}, and \ref{fig:openings_stretch} respectively.
For each opening group, experts also suggest the most likely winning move for White. 
We split these openings into several three-move openings by exploring Black's possible replies. 
For each opening group, we select the most difficult replies by Black according to expert recommendations and the PCN policy head output. 
For simplicity, in the rest of the paper, \textit{JA} represents the position resulting from Black playing at A in the jump group; \textit{KB} represents Black playing at B in the knight's move group, etc. 
The gray solid squares represent moves that are also suggested by the PCN, but cannot be solved by both of the baseline and online fine-tuning solvers in one day.
With limited computing resources, we leave these openings for future work. 
In total, we use 16 three-move openings as shown in Figure \ref{fig:openings}.

\subsection{Baseline versus online fine-tuning}
\label{sec:exp_baseline_vs_online_learning}

Table \ref{tab:baseline-online} lists statistics for solving the 16 three-move openings by the baseline solver and three variants of online fine-tuning solvers. 
In general, all online fine-tuning solvers outperform the baseline solver in most openings.
\textsc{online-sp}, \textsc{online-cp}, and \textsc{online-sp+cp}, require only about 48.53\%, 21.69\%, and 23.07\% of the visited nodes, and 52.67\%, 23.54\%, and 24.99\% of the computing time compared to \textsc{baseline}.
This shows that fine-tuning PCNs with critical positions, which are currently being solved by either the manager or workers, provides better heuristics for the current search tree and accelerates the solving process.
Furthermore, \textsc{online-sp+cp} has nearly the same performance as \textsc{online-cp}, with both methods outperforming \textsc{online-sp}.
This means that training with critical positions is more important than solved positions.
To reduce the overhead of sending both solved and critical positions, we simply choose \textsc{online-cp} for further analysis, i.e. 
all instances of \textit{online fine-tuning solver} for the rest of this section refers to \textsc{online-cp}.
In conclusion, these results indicate that $\theta_0$ provides less accurate heuristics, which impacts the proof search negatively. 
By performing online fine-tuning with either solved or critical positions, we can fine-tune the PCN dynamically according to the manager's current focus and therefore find faster solutions.

\begin{table}[h]
    \caption{The number of nodes and time to solve 16 7x7 Killall-Go three-move openings by the baseline and three variants of online fine-tuning solvers.
    ``\# Nodes'' lists the numbers of all nodes visited by the manager and workers together.
    All the listed times are rounded to the nearest second.
    The rightmost column lists the number of PCN models produced by the online fine-tuning trainer.}
    \label{tab:baseline-online}
    \vskip 0.1in
    \centering
    \begin{adjustbox}{width=\columnwidth}
    \begin{filecontents*}{tables/baseline-online.csv}
ID	Baseline Nodes	Baseline Time	Online (SP) Nodes	Online (SP) Time	Online (SP) PCN	Online (CP) Nodes	Online (CP) Time	Online (CP) PCN	Online (SP+CP) Nodes	Online (SP+CP) Time	Online(SP+CP) PCN
JA	8,964,444,959	142,115	4,054,562,593	69,699	359	\textbf{1,288,601,416}	\textbf{22,384}	186	1,425,668,707	24,865	225
JB	7,137,514,712	155,786	3,378,672,517	83,454	424	\textbf{1,576,437,139}	31,957	272	1,601,479,130	\textbf{31,455}	283
JC	721,004,784	12,514	819,264,890	13,963	57	\textbf{316,391,324}	\textbf{6,537}	59	414,108,746	8,343	69
JD	1,271,426,148	30,209	846,365,092	19,396	113	545,655,175	11,083	102	\textbf{502,966,563}	\textbf{10,896}	103
KA	134,881,952	2,103	143,814,448	2,621	14	111,838,889	2,102	18	\textbf{104,905,173}	\textbf{1,931}	18
KB	10,153,035,632	156,583	3,794,290,131	64,493	305	\textbf{2,242,789,149}	\textbf{38,947}	343	2,527,488,112	43,200	386
KC	38,217,263	747	45,217,101	1,156	6	26,441,989	758	6	\textbf{25,508,784}	\textbf{706}	6
KD	2,754,213,379	47,494	1,504,977,329	25,715	126	955,257,191	17,434	145	\textbf{920,902,808}	\textbf{16,357}	148
KE	1,197,819,407	18,771	214,614,577	3,917	21	181,418,954	3,336	30	\textbf{168,590,287}	\textbf{3,095}	28
KF	9,516,440,320	147,271	6,080,836,868	100,690	519	2,107,185,330	35,418	285	\textbf{2,027,558,505}	\textbf{35,197}	305
DA	7,322,743,383	112,874	3,015,438,589	50,046	248	1,761,842,477	30,313	266	\textbf{1,665,511,033}	\textbf{28,337}	235
SA	51,272,288	\textbf{937}	54,574,495	1,471	7	41,863,480	992	9	\textbf{41,796,555}	1,105	10
SB	215,380,103	3,860	65,970,358	1,423	7	\textbf{55,541,455}	\textbf{1,364}	12	109,591,487	2,258	20
SC	97,559,402	\textbf{1,557}	213,889,777	3,553	19	98,661,355	1,715	16	\textbf{93,535,813}	1,655	15
SD	8,187,017,679	124,644	3,821,472,453	63,058	329	\textbf{1,395,444,447}	\textbf{23,751}	154	1,485,439,307	25,531	224
SE	4,297,808,879	64,227	2,065,528,927	33,437	166	\textbf{757,256,934}	\textbf{12,465}	103	1,200,741,176	20,428	182

\end{filecontents*}
\csvreader[
        tabular={cr @{\hspace{1\tabcolsep}} rr @{\hspace{1\tabcolsep}} r @{\hspace{1\tabcolsep}} rr @{\hspace{1\tabcolsep}} r @{\hspace{1\tabcolsep}} rr @{\hspace{1\tabcolsep}} r @{\hspace{1\tabcolsep}} r},
        separator = tab,
        table head = \toprule & \multicolumn{2}{c}{\textsc{baseline}} & \multicolumn{3}{c}{\textsc{online-sp}} & \multicolumn{3}{c}{\textsc{online-cp}} & \multicolumn{3}{c}{\textsc{online-sp+cp}}  \\ \cmidrule(lr){2-3} \cmidrule(lr){4-6} \cmidrule(lr){7-9} \cmidrule(lr){10-12}
         & \# Nodes & Time (s) & \# Nodes & Time (s) & \# PCN & \# Nodes & Time (s) & \# PCN & \# Nodes & Time (s) & \# PCN \\ \midrule,
        late after last line = \\ \midrule
        \textit{sum} & {62,060,780,290} & {1,021,692} & {30,119,490,145} & {538,092} & - & \textbf{13,462,626,704} & \textbf{240,556} & - & {14,315,792,186} & {255,359} & - \\ \bottomrule,
    ]{tables/baseline-online.csv}{}{\csvcoli & \csvcolii & \csvcoliii & \csvcoliv & \csvcolv & \csvcolvi & \csvcolvii & \csvcolviii & \csvcolix & \csvcolx & \csvcolxi & \csvcolxii}
    \end{adjustbox}
\end{table}

\begin{figure}[t]
    \centering
    \begin{minipage}{0.8\textwidth}
        \input{figures/baseline-online3-bar-chart-full.tex}
    \end{minipage}
    \hspace*{1em}
    \captionsetup[subfigure]{justification=centering}
    \caption{Solving time comparison for 16 three-move 7x7 Killall-Go openings.}
    \label{fig:baseline-online-bar-chart}
\end{figure}
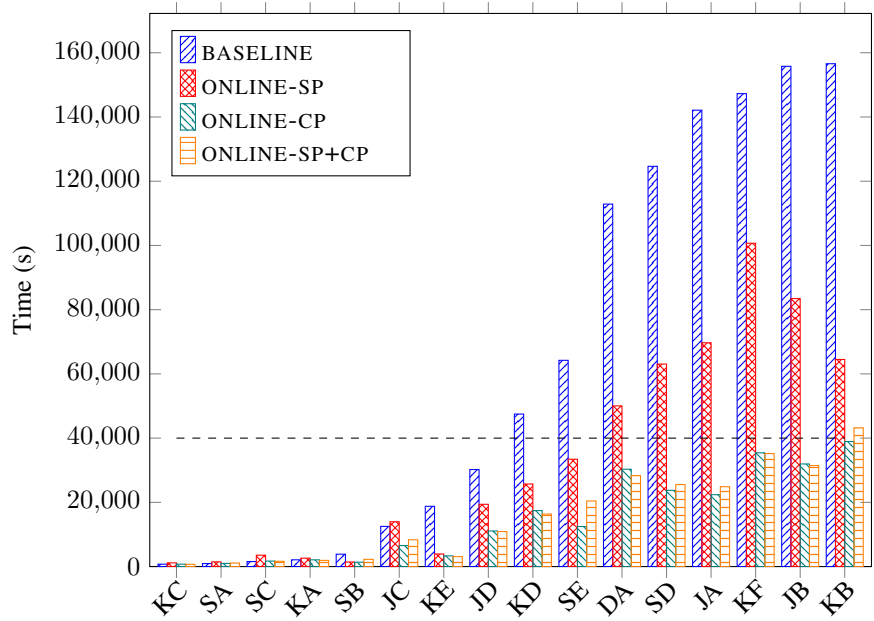

Table \ref{tab:baseline-online} also shows another interesting result: the larger the problem, the better the improvement.
For better visualization, the solving times are depicted as a bar chart in Figure \ref{fig:baseline-online-bar-chart}, where the x-axis is sorted according to the solving time of the baseline solver. 
In Figure \ref{fig:baseline-online-bar-chart}, the online fine-tuning solver solves all openings within 40,000 seconds, while the baseline solver uses more than one day to solve six openings.
Most impressively, for \textit{JA}, the online fine-tuning solver performs about 6.35 times faster than the baseline, reducing the computation time from 142,115 to 22,384 seconds, while the number of visited nodes is reduced from 8.96 billion to 1.29 billion or so nodes. 


The online fine-tuning solver does not always perform better than the baseline, especially when the openings are relatively easy to solve for the baseline. 
In Table \ref{tab:baseline-online}, the baseline uses less time to solve \textit{SA} and \textit{SC}. 
The following reasons may be why this limitation exists for smaller problems.
First, the online fine-tuning solver relies on a trainer to fine-tune the PCN. 
The quicker the problem can be solved, the less time the trainer has to fine-tune specific PCN weights. 
Consequently, it has a weaker impact on improvement. 
This is corroborated by the fact that these two openings end up with less than 20 PCN versions,\footnote{In our settings, the trainer typically generates a new PCN version about every 120 seconds.} as shown in Table \ref{tab:baseline-online}.
Second, when compared to the baseline, the online fine-tuning solver has less computing power for the workers. The trainer overhead takes up a GPU and leaves workers with seven instead of eight GPUs used by the baseline.

It is also worth mentioning that all three-move openings in both \textit{jump} and \textit{stretch} are solved. 
As these openings are considered the most difficult moves, we expect that both \textit{jump} and \textit{stretch} (i.e., the two-move openings) can probably be solved completely in the near future, with more computing resources. 

\begin{figure}[t]
    \centering
    \subfloat[Sub-position of \textit{JA}.]{
        \includegraphics[width=0.2\textwidth]{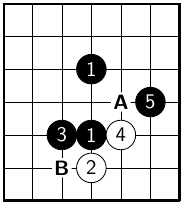}
        \label{fig:JA_different_sol}
    }
    \subfloat[Average length of critical positions.]{
        \scalebox{0.8}{
            \input{figures/position-length.tex}
        }
        \label{fig:position-length}
    }        
    \captionsetup[subfigure]{justification=centering}
    \caption{Behavioral analysis of the online fine-tuning solver for the opening JA.}
    \label{fig:move-JA}
\end{figure}

\subsection{Behavior analysis for the online fine-tuning solver}
\label{sec:exp_behavior_online}

We further analyze the behavior of two solvers by studying the opening \textit{JA}, where the online fine-tuning solver has the largest speedup among all openings, i.e. 6.35 times faster than the baseline. 
We observe several positions in which the winning moves for White differed between the two solvers. 
For example, a crucial sub-position in the solution tree is shown in Figure \ref{fig:JA_different_sol}. 
The baseline and online fine-tuning solver chose moves A and B to search, respectively.
We examine $\theta_0$ and find that the probabilities from the policy network for moves A and B are 0.416 and 0.133, respectively. 
As a result, the baseline solver has a lower chance to explore B. 
In contrast, with the help of the OFT, the online fine-tuning solver quickly realizes that solving B is faster, though it still attempted to search A initially, as it starts with the same $\theta_0$.  
In terms of node counts, the baseline spent a total of 1.63 billion nodes, with approximately 1.48 billion (91.00\%) and 35.50 million (2.18\%) nodes spent on searching A and B, respectively, while the online fine-tuning solver spent a total of 136.47 million nodes on that position, with approximately 22.13 million (16.21\%) and 95.17 million (69.74\%) nodes spent on searching A and B, respectively.
This example clearly demonstrates the advantage of using the OFT. 

Next, we investigate the set of critical positions maintained by the online fine-tuning trainer, as described in subsection \ref{sec:method_online_trainer_cp}.
During fine-tuning, the trainer randomly selects positions from the queue, then runs self-play games from these positions. 
Figure \ref{fig:position-length} shows the average path length of critical positions during training in the y-axis, and the training iteration in the x-axis, where the trainer generates a new PCN version for every iteration. 
The length starts at around 10 and gradually increases to nearly 25 in the end. 
This is because as the manager search tree grows through node expansion, the critical positions are chosen from deeper parts of the tree. 
The curve also fluctuates as the proof search progresses.
This is because the manager tends to focus on a sub-problem at a time. When a subtree is solved, the manager may then shift its attention to other unsolved parts of the proof search, which can have a relatively shallower depth. 
We also analyze similar figures for other openings in the appendix.

\subsection{Updating PCNs in online fine-tuning}
\label{sec:exp_load_model}

We investigate the impact of updating PCNs for the manager and workers during online fine-tuning. 
We select four openings, \textit{JC}, \textit{KE}, \textit{DA}, and \textit{SE}, one from each opening group, for this experiment. 
Table \ref{tab:ablation-online} summarizes results, where \textsc{baseline} denotes the baseline solver, \textsc{online-cp} denotes the online fine-tuning solver that updates PCNs for both the manager and workers (as described in subsection \ref{sec:method_online_trainer_cp}), \textsc{online-cp-m} denotes updating the PCN for the manager only, and \textsc{online-cp-w} for workers only.

First, updating PCNs for both the manager and workers performs the best.
By using consistent PCNs, jobs assigned by the manager are efficiently solved by the workers. 
With inconsistent PCNs, the results can be even worse than the baseline. 
Generally, \textsc{online-cp-m} outperforms \textsc{online-cp-w}, except for the opening \textit{SE}.
We find that the pre-trained PCN causes \textsc{online-cp-w} to divide its computing resources across several OR nodes (white moves), many of which are relatively difficult to solve. 
In contrast, for \textsc{online-cp-m}, with the updated PCNs, the manager focuses on one white move (or a smaller number of white moves), of which they are much easier to solve. 
Thus, even if there is a mismatch between the manager and workers in \textsc{online-cp-m}, focusing on one good white move can still result in efficiency.

\begin{table}[h]
  \centering
  \begin{minipage}[t]{0.44\linewidth}
    \centering
    \caption{Impact of updating PCNs in the online fine-tuning solver. The number represents the time (in seconds) for solving each opening by using different methods.}
    \label{tab:ablation-online}
    \vskip 0.1in
    \begin{adjustbox}{width=\columnwidth}
    \begin{tabular}{lrrrr}
    \toprule
    & JC & KE & DA & SE \\
    \midrule
    \textsc{baseline} & 12,514 & 18,771 & 112,874 & 64,227 \\
    \textsc{online-cp-m} & 9,992 & 5,445 & 46,719 & 28,769 \\
    \textsc{online-cp-w} & 13,516 & 54,360 & 65,304 & 26,168 \\
    \textsc{online-cp} & \textbf{6,537} & \textbf{3,336} & \textbf{30,313} & \textbf{12,465} \\
    \bottomrule
    \end{tabular}
    \end{adjustbox}
  \end{minipage}
  \hspace*{0.1em}
  \begin{minipage}[t]{0.54\linewidth}
    \centering
    \caption{Ablation study for job assignment schemes in the online fine-tuning solver. The number represents the time (in seconds) for solving each opening by using different methods.}
    \label{tab:ablation-manager}
    \vskip 0.1in
    \begin{adjustbox}{width=\columnwidth}
    \begin{tabular}{lrrrr}
    \toprule
    & JC & KE & DA & SE \\
    \midrule
    \textsc{online-cp} & 6,537 & \textbf{3,336} & \textbf{30,313} & \textbf{12,465} \\
    \ w/o AND assg. & 8,618 & 3,762 & 53,275 & 18,926 \\
    \ w/o top-\textit{k} & \textbf{6,520} & 3,639 & 46,080 & 36,093 \\
    \ w/o AND assg. \& top-\textit{k} & 10,480 & 6,171 & 58,527 & 33,677 \\
    \bottomrule
    \end{tabular}
    \end{adjustbox}
  \end{minipage}
\end{table}

\subsection{Ablation study for job assignment schemes}
\label{sec:exp_ablation_study}

We conduct an ablation study in the online fine-tuning solver to analyze the impact of job assignment schemes, described in subsection \ref{sec:method_job_assignment}.
We only include ablations for top-$k$ selection and AND-player job assignment, since virtual solving is required to avoid job redundancy.
The ablation study is performed on the same four openings as subsection \ref{sec:exp_load_model}, \textit{JC}, \textit{KE}, \textit{DA}, and \textit{SE}.
Table \ref{tab:ablation-manager} summarizes the ablation results, where \textsc{online-cp} denotes the online fine-tuning solver that uses both schemes; the other three versions denote the ablations by removing specific schemes from \textsc{online-cp}.
If the top-$k$ selection is removed, the manager always selects the best child during selection for both AND-nodes and OR-nodes.
If we do not follow the AND-player job assignment scheme (abbreviated as \textit{AND assg.} in the table), the manager assigns both AND-player and OR-player jobs.

From Table \ref{tab:ablation-manager}, \textsc{online-cp} performs the best in general.
In particular, \textsc{online-cp} only requires around 48.37\% of the computing time on average over all four openings, compared to the solver without both schemes (the last row in the table).
When comparing each technique individually, the improvement varies from problem to problem. 



\section{Discussion}
\label{sec:discussion}
This paper demonstrates the potential of using online fine-tuning for game solving.
On average across multiple openings, our proposed online fine-tuning solver only uses 23.54\% of the computation time compared to the baseline.
Our distributed game solver is the first online fine-tuning method for problem solving based on AlphaZero-like algorithms.
Although we focus on online fine-tuning throughout this paper, we can also claim that the complete distributed game solver is a life-long learning system.
The online trainer continuously refines heuristics for unfamiliar (unsolved) positions, and forgets the previously learned knowledge (solved positions).
However, this forgotten knowledge is saved (remembered) in the manager's solution tree. As a result, the worker and trainer will not need to evaluate these positions again.

There are many other topics for future investigation.
Our experiments on the four challenging 7x7 Killall-Go opening groups show that two groups are likely to be solved in the near future.
However, for the other two, or even 7x7 Killall-Go in its entirety, we expect more novel techniques are needed. 
As for the standard Go game, the largest solved board to date is only 5x6 in size \cite{van2009solving}, with no published progress in 14 years.
We expect online fine-tuning to be one of the key improvements that can help push this boundary.
As for generality, our method is not limited to Go but can be easily applied to other two-player zero-sum games like Hex or Othello.
Moreover, we expect it has the potential to extend to single-player games such as Rubik's Cube, or even to other non-game fields, such as automated theorem proving \cite{lample2022hypertree} or chemical syntheses \cite{segler2018planning, Kishimoto2019DepthFirstPS}.

\begin{ack}
This research is partially supported by the National Science and Technology Council (NSTC) of the Republic of China (Taiwan) under Grant Number NSTC 111-2222-E-001-001-MY2, NSTC 111-2221-E-A49-101-MY2, NSTC 110-2221-E-A49-067-MY3, and NSTC 111-2634-F-A49-013.
\end{ack}

%% file: figures/baseline-online3-bar-chart-full.tex
\begin{tikzpicture}[scale=1]
\begin{axis}[
    ybar = 0pt,
    legend pos = north west,
    legend columns = 1,
    legend style = {text width=2.6cm, align=left},
    compat = newest,
    ylabel = {Time (s)},
    width = \columnwidth,
    height = 0.8\columnwidth,
    ymin = 0,
    xtick = data,
    xticklabels from table = {figures/baseline-online-bar-chart.csv}{ID},
    x tick label style = {rotate=45, anchor=north east, inner sep=0mm},
    ytick = {0, 20000, 40000, 60000, 80000, 100000, 120000, 140000, 160000},
    yticklabel style={/pgf/number format/fixed},
    scaled y ticks = false,
    enlarge x limits = 0.04,
    bar width = 3.5pt,
    legend image code/.code={\draw [#1] (0cm,-0.1cm) rectangle (0.2cm,0.25cm); },
]

\addplot+[ybar, pattern=north east lines, pattern color=blue] table [
    x expr=\coordindex,
    y = Baseline,
    col sep = tab,
] {figures/baseline-online-bar-chart.csv};
\addlegendentry{\textsc{baseline}}

\addplot+[ybar, pattern=crosshatch, pattern color=red] table [
    x expr=\coordindex,
    y = Online(SP),
    col sep = tab,
] {figures/baseline-online-bar-chart.csv};
\addlegendentry{\textsc{online-sp}}

\addplot+[ybar, color=teal, pattern=north west lines, pattern color=teal] table [
    x expr=\coordindex,
    y = Online(CP),
    col sep = tab,
] {figures/baseline-online-bar-chart.csv};
\addlegendentry{\textsc{online-cp}}

\addplot+[ybar, color=orange, pattern=horizontal lines, pattern color=orange] table [
    x expr=\coordindex,
    y = Online(SP+CP),
    col sep = tab,
] {figures/baseline-online-bar-chart.csv};
\addlegendentry{\textsc{online-sp+cp}}

\addplot [
    black,
    dashed,
    sharp plot,
] coordinates {(0,40000) (15,40000)};

\end{axis}
\end{tikzpicture}

\nointerlineskip

%% file: figures/position-length.tex
\begin{tikzpicture}[scale=0.72]
\begin{axis} [
    xlabel = Iteration,
    ylabel = Length,
    width = \columnwidth,
    height = 0.5\columnwidth,
    compat = newest,
    grid style = dashed,
    ymajorgrids = true,
    enlarge x limits = 0.02,
    tick label style = {font=\scriptsize},
    label style = {font=\scriptsize},
    mark size = 1pt,
]
\addplot+ [
    mark = x,
] table [
    x = Iteration,
    y = JA,
    col sep = tab
] {figures/position-length.csv};

\end{axis}
\end{tikzpicture}

%% file: appendix.tex
\appendix
\section{Implementation details}
\label{sec:appendix_implementaion}

\subsection{PCN training}
\label{sec:appendix_pcn_training}
We basically follow the same PCN training method by \citet{wu2022alphazero} but replace the AlphaZero algorithm with the Gumbel AlphaZero algorithm \cite{danihelka2021policy}, where the simulation count is set to 32\footnote{The original PCN training used 400 simulation counts in the self-play, requiring much more computing resources than using Gumbel algorithm.} in self-play and starts by sampling 16 actions.
The architecture of the PCN contains three residual blocks with 256 hidden channels.
A total of 400,000 self-play games are generated for the whole training.
During optimization, the learning rate is fixed at 0.02, and the batch size is set to 1,024.
The PCN is optimized for 500 steps for every 2,000 self-play games.
The pre-trained PCN requires around 13 hours to train on a machine with four 1080Ti GPUs, i.e. 52 1080Ti GPU-hours.
For the online trainer, we use the same hyperparameters as the pre-trained PCN but only use one GPU.

\subsection{7x7 Killall-Go solver}
\label{sec:appendix_solver}
Our solver is built upon the state-of-the-art (SOTA) 7x7 Killall-Go solver \cite{shih2022novel} except for the following three changes.
First, our solver uses PCN as heuristics while the SOTA solver trains a network with \textit{Faster to Life} (FTL) techniques.
Both networks aim to provide a faster move for solving, but FTL requires additional (\textit{komi}\footnote{Since Black plays the first stone in the game of Go, White usually earns some extra points called komi for balance.}) settings in solving, so PCN is much easier to use in our solver.
Second, we implement the transposition table based on \citet{shih2023local}.
This greatly reduces the solving time.
Finally, we implement a solution for resolving \textit{Graph-History-Interaction} (GHI, i.e. cycles in Go) \cite{palay1983searching} problems to ensure the correctness of reusing solutions in the transposition table, based on \citet{kishimoto2004general, kishimoto2005solution}'s GHI solution.

\subsection{Worker design}
\label{sec:appendix_worker}
The worker is itself a Killall-Go solver. It is GPU bound, i.e. it relies on GPUs more than CPUs since the PCN (a neural network) requires intensive GPU computation.
Thus, to fully utilize GPU resources, we implement batch GPU inferencing to accelerate PCN evaluations for workers.
In practice, we collect 48 workers together in one process with multiple threads.
The process runs MCTS selection for each worker independently.
Namely, a total of 48 leaf nodes are generated and evaluated by PCN with one GPU at once.
The 48 leaf nodes are collected as a batch for batch GPU inferencing, with a batch size of 48.
This method greatly reduces the solving time when more workers are used.
The baseline distributed game solver creates eight processes as workers, each with one GPU, for a total of 384 workers (eight processes with 48 workers).
The online fine-tuning solver has the same number of workers for fairness, but uses seven GPUs (one GPU is spared for the online trainer); the configuration is six processes with 55 workers and one process with 54 workers.


\section{Experiment details}
\label{sec:appendix_exp}

\subsection{Setup}
\label{sec:appendix_exp_setup}

All experiments are conducted in three machines, each equipped with two Intel Xeon E5-2678 v3 CPUs, 192G RAM, and four GTX 1080Ti GPUs. 
We list other hyperparameters in Table \ref{tab:hyperparameters}.

For the memory used in solving, the manager requires 20G RAM for expanding every 1M nodes, and every 48 workers together in one process requires 30G RAM at most.
Note that workers use the same amount of memory regardless of problem size.
They are limited to 100,000 nodes per job; the job result is “unsolved” if a solution is not obtained within that limit.

Specifically, for \textsc{baseline} with 384 workers, solving \textit{KA} used 2,103 seconds, required 3G RAM for the manager and 240G RAM for the workers; solving \textit{KB} used 156,583 seconds, required 170G RAM for the manager and 240G RAM for the workers.
However, for \textsc{baseline} with only 48 workers, solving \textit{KA} used 12,151 seconds but only required 2G RAM for the manager and 30G RAM for the workers.
Overall, the settings can be varied depending on available machines.

\begin{table}[h]
    \caption{Hyperparameters used in the baseline and online fine-tuning solvers. All variants of online fine-tuning solvers use the same settings.}
    \label{tab:hyperparameters}
    \vskip 0.1in
    \centering
    \begin{tabular}{crcc}
        \toprule
        & & \textsc{baseline} & \textsc{online} \\ \midrule
        \multirowcell{4}{\textbf{Manager}}
        & \# GPUs & 1 & 1 \\ 
        & $v_{thr}$ & 16.5 & 16.5 \\ 
        & $k$ for top-k selection & 4 & 4 \\ \midrule 
        \multirowcell{3}{\textbf{Worker}}
        & \# GPUs & 8 & 7 \\
        & \# workers & 384 & 384 \\ 
        & \# node limitation per job & 100,000 & 100,000 \\ \midrule 
        \multirowcell{1}{\textbf{Trainer}}
        & \# GPUs & 0 & 1 \\
        \bottomrule
    \end{tabular}
\end{table}

\subsection{Scalability of the distributed game solver}
\label{sec:exp_scaling_performance}

To evaluate the scalability of the distributed game solver, we run \textsc{baseline} with different numbers of workers on \textit{KA}. Specifically, the solvers use 384, 192, 96, and 48 workers, using 8, 4, 2, and 1 GPU, respectively. Every 48 workers share one GPU. The results are shown in Table \ref{tab:scaleup}. Overall, the speedup is around 1.8 times faster when the number of workers is doubled (up to 384 workers due to our machine limitation).

\begin{table}[h]
    \caption{Detailed statistics for solving \textit{KA} by \textsc{baseline} with different numbers of workers.}
    \label{tab:scaleup}
    \vskip 0.1in
    \centering
    \begin{adjustbox}{width=\columnwidth}
    \begin{filecontents*}{tables/scaleup.csv}
384	134,881,952	2,103	121,236	21,748	34.48	6,196.46	0	97.87%	94.53%	5.78
192	120,676,465	3,596	99,678	18,598	35.92	6,483.32	0	98.44%	98.57%	3.09
96	112,344,894	6,502	84,752	16,422	37.45	6,835.96	0	98.87%	98.90%	1.71
48	109,362,406	12,151	74,665	15,292	37.78	7,146.73	0	99.05%	98.60%	1.00
\end{filecontents*}
\csvreader[
        tabular={c r @{\hspace{1\tabcolsep}} r r @{\hspace{1\tabcolsep}} r r @{\hspace{1\tabcolsep}} r r @{\hspace{1\tabcolsep}} r r r},
        separator=tab,
        table head=\toprule \multirowcell{2}{\# Workers} & \multirowcell{2}{\# Nodes} & \multirowcell{2}{Time (s)} & \multirowcell{2}{Manager\\\# Nodes} & \multirowcell{2}{\# Jobs} & \multirowcell{2}{Avg. Job\\Time (s)} & \multirowcell{2}{Avg. Job \\\# Nodes} & \multirowcell{2}{\# PCN} & \multirowcell{2}{Solved\\Jobs (\%)} & \multirowcell{2}{Avg. Worker\\Loading (\%)} & \multirowcell{2}{Speedup} \\
        & & & & & & & & &
        \\ \midrule,
        late after last line=\\ \bottomrule,
        no head,
        respect all
    ]{tables/scaleup.csv}{}{\csvcoli & \csvcolii & \csvcoliii & \csvcoliv & \csvcolv & \csvcolvi & \csvcolvii & \csvcolviii & \csvcolix & \csvcolx & \csvcolxi}
    \end{adjustbox}
\end{table}

\subsection{Statistics of solving 7x7 Killall-Go three-move openings}
\label{sec:appendix_exp_solution}
Figure \ref{fig:winning_moves} shows the next winning moves (the fourth moves) of 16 three-move openings for both baseline and \textsc{online-cp} solvers.
Generally, both solvers solve the openings at the same next moves, except \textit{JB}.
The full solution trees for each opening can be found in this link: \url{https://rlg.iis.sinica.edu.tw/papers/neurips2023-online-fine-tuning-solver/solution-trees}.
We also provide a tool and a README file for explaining the solution tree.

\input{figures/sgfs.tex}

It is worth mentioning that \textit{JA} and \textit{JB} are similar to one of the common josekis\footnote{A joseki is a move sequence that is widely believed to be balanced play by both players.} played in 19x19 Go.
The joseki usually occurs when Black makes a \textit{corner enclosure} move, also known as \textit{shimari} in Japanese, like the two stones marked as ``1'' in \textit{JA} and \textit{JB}.
Then, White attempts to invade Black's territories by playing at the stone marked as ``2''.
Judging by the online fine-tuning solver's ability to solve \textit{JA} and \textit{JB}, we foresee a high potential to extend our work to solving other 19x19 Go corner josekis in the future.

In addition, Figure \ref{fig:position-length-detailed} shows the curve for average critical position lengths. 
These curves are all similar in the sense that it starts with small average lengths, which gradually increases during fine-tuning.

\input{figures/detailed-position-length.tex}

Table \ref{tab:detailed-baseline}, Table \ref{tab:detailed-online-ts}, Table \ref{tab:detailed-online-as} and Table \ref{tab:detailed-online} list the experiment results of the baseline and three variants of online fine-tuning solvers respectively, in more detail than those in Table 1 in the main text. 
These tables include the number of nodes for solving, the solving time in seconds, the number of nodes used in the manager, the number of jobs, the average time for solving each job, the average number of nodes for solving each job, the number of updated PCNs, the success rate of solving jobs, and the average worker load during solving.
In general, the solving time is correlated with the number of nodes and the number of jobs. 
For online fine-tuning, the solving time is also correlated with the number of PCNs as the trainer updates PCNs at a stable speed. 
Note that the number of PCNs is always $0$ for the baseline solver, as they do not update PCNs during solving.

In our experiments, the average success rates of solving jobs are around 97.30\%, 98.44\%, 99.14\% and 99.08\% for the baseline and the online fine-tuning solvers, respectively. 
In addition, for some quickly solved openings, e.g. \textit{KC}, \textit{SA}, and \textit{SB}, the average time for solving each job is far less than other difficult openings.
While the workers are able to solve jobs quickly, the managers are relatively unable to create enough jobs for the workers, causing the workers to be relatively idle (lower avg. worker loading).
Compared with the baseline solver, online fine-tuning solvers have better success rates of solving as well as lesser nodes for each job. 
This confirms that online fine-tuning successfully fine-tuned the PCNs for critical positions that the manager is interested in, thereby increasing the job efficiency overall.

\begin{table}[!h]
    \caption{Detailed statistics for the openings solved by \textsc{baseline}.}
    \label{tab:detailed-baseline}
    \vskip 0.1in
    \centering
    \begin{adjustbox}{width=\columnwidth}
    \begin{small}
        \begin{filecontents*}{tables/detailed-baseline.csv}
ID	nodes	time	count	jobs	avg time	avg nodes	load model	solved rate	avg loading
JA	8,964,444,959	142,115	4,842,554	792,465	68.68	11,305.99	0	96.83%	99.48%
JB	7,137,514,712	155,786	3,689,548	635,263	93.90	11,229.72	0	96.80%	99.48%
JC	721,004,784	12,514	900,221	165,308	23.66	4,356.14	0	98.96%	73.70%
JD	1,271,426,148	30,209	655,078	128,885	89.32	9,859.73	0	97.75%	98.96%
KA	134,881,952	2,103	121,236	21,748	34.48	6,196.46	0	97.87%	94.53%
KB	10,153,035,632	156,583	8,241,207	1,240,258	48.34	8,179.58	0	98.20%	99.48%
KC	38,217,263	747	72,880	15,284	10.33	2,495.71	0	98.39%	62.02%
KD	2,754,213,379	47,494	1,499,735	246,500	73.67	11,167.20	0	97.07%	99.22%
KE	1,197,819,407	18,771	1,024,660	150,490	47.14	7,952.65	0	97.79%	98.44%
KF	9,516,440,320	147,271	5,208,724	789,225	71.50	12,051.36	0	96.88%	99.68%
DA	7,322,743,383	112,874	4,326,195	636,200	67.90	11,503.33	0	95.62%	99.22%
SA	51,272,288	937	79,967	17,772	14.26	2,880.50	0	98.37%	75.78%
SB	215,380,103	3,860	288,751	52,191	22.91	4,121.23	0	97.99%	78.65%
SC	97,559,402	1,557	113,821	22,376	23.31	4,354.92	0	97.94%	90.62%
SD	8,187,017,679	124,644	4,286,025	668,654	71.36	12,237.62	0	95.18%	99.48%
SE	4,297,808,879	64,227	2,234,093	345,124	71.00	12,446.47	0	95.10%	98.86%
\end{filecontents*}
\csvreader[
            tabular={crrrrrrrrr},
            separator = tab,
            table head = \toprule & \multirowcell{2}{\# Nodes} & \multirowcell{2}{Time (s)} & \multirowcell{2}{Manager\\\# Nodes} & \multirowcell{2}{\# Jobs} & \multirowcell{2}{Avg. Job\\Time (s)} & \multirowcell{2}{Avg. Job \\\# Nodes} & \multirowcell{2}{\# PCN} & \multirowcell{2}{Solved\\Jobs (\%)} & \multirowcell{2}{Avg. Worker\\Loading (\%)} \\
            & & & & & & & &
            \\ \midrule,
            late after last line = \\ \bottomrule,
        ]{tables/detailed-baseline.csv}{}{\csvcoli & \csvcolii & \csvcoliii & \csvcoliv & \csvcolv & \csvcolvi & \csvcolvii & \csvcolviii & \csvcolix & \csvcolx}
    \end{small}
    \end{adjustbox}
\end{table}

\begin{table}[!h]
    \caption{Detailed statistics for the openings solved by \textsc{online-sp}.}
    \label{tab:detailed-online-ts}
    \vskip 0.1in
    \centering
    \begin{adjustbox}{width=\columnwidth}
    \begin{small}
        \begin{filecontents*}{tables/detailed-online-SP.csv}
ID	nodes	time	count	jobs	avg time	avg nodes	load model	solved rate	avg loading
JA	4,054,562,593	69,699	2,491,326	479,455	55.45	8,451.41	359	98.49%	98.95%
JB	3,378,672,517	83,454	1,607,080	327,626	97.49	10,307.68	424	97.27%	99.14%
JC	819,264,890	13,963	704,540	137,752	35.96	5,942.28	57	98.08%	88.34%
JD	846,365,092	19,396	500,259	106,495	69.00	7,942.77	113	98.50%	98.91%
KA	143,814,448	2,621	174,202	34,222	26.17	4,197.31	14	98.55%	90.25%
KB	3,794,290,131	64,493	3,671,889	548,306	43.54	6,913.33	305	98.99%	99.05%
KC	45,217,101	1,156	100,985	21,847	11.12	2,065.09	6	99.47%	54.07%
KD	1,504,977,329	25,715	986,849	202,651	48.37	7,421.58	126	98.86%	98.95%
KE	214,614,577	3,917	246,259	50,422	28.29	4,251.48	21	98.85%	95.48%
KF	6,080,836,868	100,690	5,431,753	855,725	44.93	7,099.72	519	99.19%	99.21%
DA	3,015,438,589	50,046	2,682,998	418,088	45.56	7,206.03	248	98.55%	98.90%
SA	54,574,495	1,471	122,611	25,403	11.78	2,143.52	7	98.47%	56.24%
SB	65,970,358	1,423	124,986	25,917	15.46	2,540.62	7	98.10%	79.84%
SC	213,889,777	3,553	141,447	32,739	39.32	6,528.86	19	98.06%	95.22%
SD	3,821,472,453	63,058	2,224,352	406,191	59.27	9,402.59	329	97.37%	99.01%
SE	2,065,528,927	33,437	1,647,455	282,992	44.79	7,293.07	166	98.21%	98.26%
\end{filecontents*}
\csvreader[
            tabular={crrrrrrrrr},
            separator = tab,
            table head = \toprule & \multirowcell{2}{\# Nodes} & \multirowcell{2}{Time (s)} & \multirowcell{2}{Manager\\\# Nodes} & \multirowcell{2}{\# Jobs} & \multirowcell{2}{Avg. Job\\Time (s)} & \multirowcell{2}{Avg. Job \\\# Nodes} & \multirowcell{2}{\# PCN} & \multirowcell{2}{Solved\\Jobs (\%)} & \multirowcell{2}{Avg. Worker\\Loading (\%)} \\
            & & & & & & & &
            \\ \midrule,
            late after last line = \\ \bottomrule,
        ]{tables/detailed-online-SP.csv}{}{\csvcoli & \csvcolii & \csvcoliii & \csvcoliv & \csvcolv & \csvcolvi & \csvcolvii & \csvcolviii & \csvcolix & \csvcolx}
    \end{small}
    \end{adjustbox}
\end{table}

\begin{table}[!h]
    \caption{Detailed statistics for the openings solved by \textsc{online-cp}.}
    \label{tab:detailed-online-as}
    \vskip 0.1in
    \centering
    \begin{adjustbox}{width=\columnwidth}
    \begin{small}
        \begin{filecontents*}{tables/detailed-online-CP.csv}
ID	nodes	time	count	jobs	avg time	avg nodes	load model	solved rate	avg loading
JA	1,288,601,416	22,384	1,314,785	259,008	32.68	4,970.07	186	99.48%	98.44%
JB	1,576,437,139	31,957	1,643,806	327,442	36.67	4,809.38	272	99.46%	96.66%
JC	316,391,324	6,537	538,369	109,298	17.09	2,889.83	59	99.27%	72.38%
JD	545,655,175	11,083	501,953	109,891	38.12	4,960.85	102	99.32%	98.75%
KA	111,838,889	2,102	159,614	32,153	22.60	3,473.37	18	98.70%	92.69%
KB	2,242,789,149	38,947	3,202,296	575,365	24.45	3,892.46	343	99.65%	93.25%
KC	26,441,989	758	69,052	15,423	9.12	1,709.97	6	99.01%	50.78%
KD	955,257,191	17,434	1,222,772	227,427	26.86	4,194.90	145	99.69%	89.35%
KE	181,418,954	3,336	261,253	49,484	24.25	3,660.93	30	98.86%	94.53%
KF	2,107,185,330	35,418	2,555,399	427,993	31.27	4,917.44	285	99.60%	98.06%
DA	1,761,842,477	30,313	2,556,573	421,268	26.16	4,176.17	266	99.60%	94.03%
SA	41,863,480	992	86,749	19,634	12.60	2,127.77	9	98.57%	70.96%
SB	55,541,455	1,364	125,338	27,612	11.64	2,006.96	12	98.48%	64.32%
SC	98,661,355	1,715	116,980	24,770	24.34	3,978.38	16	98.17%	94.28%
SD	1,395,444,447	23,751	1,575,572	278,198	32.24	5,010.35	195	99.13%	98.33%
SE	757,256,934	12,465	892,615	153,343	30.43	4,932.50	103	99.17%	98.26%
\end{filecontents*}
\csvreader[
            tabular={crrrrrrrrr},
            separator = tab,
            table head = \toprule & \multirowcell{2}{\# Nodes} & \multirowcell{2}{Time (s)} & \multirowcell{2}{Manager\\\# Nodes} & \multirowcell{2}{\# Jobs} & \multirowcell{2}{Avg. Job\\Time (s)} & \multirowcell{2}{Avg. Job \\\# Nodes} & \multirowcell{2}{\# PCN} & \multirowcell{2}{Solved\\Jobs (\%)} & \multirowcell{2}{Avg. Worker\\Loading (\%)} \\
            & & & & & & & &
            \\ \midrule,
            late after last line = \\ \bottomrule,
        ]{tables/detailed-online-CP.csv}{}{\csvcoli & \csvcolii & \csvcoliii & \csvcoliv & \csvcolv & \csvcolvi & \csvcolvii & \csvcolviii & \csvcolix & \csvcolx}
    \end{small}
    \end{adjustbox}
\end{table}

\begin{table}[!h]
    \caption{Detailed statistics for the openings solved by \textsc{online-sp+cp}.}
    \label{tab:detailed-online}
    \vskip 0.1in
    \centering
    \begin{adjustbox}{width=\columnwidth}
    \begin{small}
        \begin{filecontents*}{tables/detailed-online-SP+CP.csv}
ID	nodes	time	count	jobs	avg time	avg nodes	load model	solved rate	avg loading
JA	1,425,668,707	24,865	1,370,665	278,183	33.85	5,120.00	225	99.45%	98.18%
JB	1,601,479,130	31,455	1,743,905	351,934	33.16	4,545.55	283	99.50%	95.31%
JC	414,108,746	8,343	693,560	140,608	17.68	2,940.20	69	99.54%	75.26%
JD	502,966,563	10,896	401,057	89,954	45.82	5,586.92	103	99.13%	98.44%
KA	104,905,173	1,931	109,111	23,739	28.97	4,414.51	18	98.21%	95.83%
KB	2,527,488,112	43,200	3,148,579	578,009	27.79	4,367.30	386	99.59%	96.09%
KC	25,508,784	706	58,399	13,172	10.58	1,932.16	6	99.18%	53.65%
KD	920,902,808	16,357	1,092,352	210,345	27.80	4,372.87	148	99.57%	91.15%
KE	168,590,287	3,095	214,173	42,447	26.24	3,966.74	28	98.81%	94.79%
KF	2,027,558,505	35,197	2,203,830	383,317	34.79	5,283.76	305	99.57%	98.33%
DA	1,665,511,033	28,337	2,252,356	377,189	27.77	4,409.62	235	99.53%	95.57%
SA	41,796,555	1,105	95,672	21,325	10.67	1,955.49	10	98.67%	55.73%
SB	109,591,487	2,258	167,238	34,085	19.62	3,210.33	20	98.04%	78.12%
SC	93,535,813	1,655	94,822	21,820	26.28	4,282.36	15	98.12%	93.49%
SD	1,485,439,307	25,531	1,674,274	296,617	32.26	5,002.29	224	99.15%	97.12%
SE	1,200,741,176	20,428	1,289,405	231,498	33.15	5,181.26	182	99.17%	98.01%
\end{filecontents*}
\csvreader[
            tabular={crrrrrrrrr},
            separator = tab,
            table head = \toprule & \multirowcell{2}{\# Nodes} & \multirowcell{2}{Time (s)} & \multirowcell{2}{Manager\\\# Nodes} & \multirowcell{2}{\# Jobs} & \multirowcell{2}{Avg. Job\\Time (s)} & \multirowcell{2}{Avg. Job \\\# Nodes} & \multirowcell{2}{\# PCN} & \multirowcell{2}{Solved\\Jobs (\%)} & \multirowcell{2}{Avg. Worker\\Loading (\%)} \\
            & & & & & & & &
            \\ \midrule,
            late after last line = \\ \bottomrule,
        ]{tables/detailed-online-SP+CP.csv}{}{\csvcoli & \csvcolii & \csvcoliii & \csvcoliv & \csvcolv & \csvcolvi & \csvcolvii & \csvcolviii & \csvcolix & \csvcolx}
    \end{small}
    \end{adjustbox}
\end{table}

\subsection{Different PCN thresholds}
\label{sec:appendix_PCN_threshold}

We examine different $v_{thr}$ from $11.5$ to $21.5$ on opening \textit{JC}, using the baseline solver. 
The experiment result is presented in Table \ref{tab:pcn-threshold}, where the four columns represent the examined $v_{thr}$, the total solving time, the average time for workers to solve jobs, and the job success rate. 
Among these PCN thresholds, we consider $v_{thr} = 16.5$ to be a balanced setting as it performs well in the three metrics. 
However, the results also show that the performance is not necessarily sensitive to different $v_{thr}$ settings, i.e. the solving time is similar when $v_{thr}\in(15.5,17.5)$.

As demonstrated in the table, $v_{thr}$ outside of this range deteriorates the solving performance. 
On the one hand, when $v_{thr}$ is too high, e.g. $v_{thr} = 21.5$, only about 95\% of jobs can be solved, implying that about 5\% of the jobs are wasted. 
On the other hand, when $v_{thr}$ is too low, e.g. $v_{thr} = 11.5$, the assigned jobs can be solved quickly with a high success rate. 
However, this requires the manager to assign more jobs, which increases the overhead of handling job assignments between the manager and the workers, thereby increasing the solving time. 
Note that the appropriate $v_{thr}$ may vary for different games and for different numbers of available workers. 
It is possible to adjust $v_{thr}$ dynamically during solving, which is left for future work.

\begin{table}[h]
    \caption{The solving time, average job completion time, and success rate of solvable jobs for solving opening \textit{JC} by the baseline solver with different PCN thresholds.}
    \label{tab:pcn-threshold}
    \vskip 0.1in
    \centering
    \begin{small}
        \begin{filecontents*}{tables/pcn-threshold.csv}
v_thr	Time (s)	Jobs Time (s)	Solved Jobs (%)
11.5	23,559	2.00	99.92
12.5	22,870	3.60	99.84
13.5	18,356	5.75	99.74
14.5	19,458	12.06	99.55
15.5	12,519	16.29	99.29
16.5	12,514	23.66	98.96
17.5	12,877	33.73	98.34
18.5	17,536	46.06	97.35
19.5	22,343	52.04	96.75
20.5	24,469	58.73	95.99
21.5	27,810	70.50	94.94
\end{filecontents*}
\csvreader[
            tabular = {rrrr},
            separator = tab,
            table head = \toprule $v_{thr}$ & Time (s) & Avg. Job Time (s) & Solved Jobs (\%) \\ \midrule,
            late after last line = \\ \bottomrule,
        ]{tables/pcn-threshold.csv}{}{
        \csvcoli & \csvcolii & \csvcoliii & \csvcoliv \%}
    \end{small}
\end{table}

\subsection{Comparison to offline fine-tuning}
\label{sec:exp_offline_fine_tuning}

We now investigate how much benefit we can gain from offline fine-tuning for a specific opening.
To do this, we first train $\theta_0$ by generating 400,000 self-play games (around 52 1080Ti GPU-hours) from the empty board.
The resulting network is the same as the one referred to as $\theta_0$ in the main text.
Next, we fine-tune $\theta_0$ by generating 200,000 additional self-play games (around 26 1080Ti GPU-hours) from the specific opening we are interested in. That is, if we want to solve the opening \textit{JC}, we generate self-play games starting from that opening, and perform updates on $\theta_0$ to obtain what we refer to as $\theta^{'}_0$-\textit{JC}.
For this experiment, we used four openings, so the networks $\theta^{'}_0$-\textit{JC}, $\theta^{'}_0$-\textit{KE}, $\theta^{'}_0$-\textit{DA}, and $\theta^{'}_0$-\textit{SE}
were produced.
Lastly, in the baseline case, we do not update the network with critical positions; the same network is used all throughout the proof search. In \textsc{online-cp}, critical positions are chosen and the $\theta^{'}_0$ is further fine-tuned using the OFT (resulting in 
$\theta^{'}_1$, $\theta^{'}_2$, ..., $\theta^{'}_t$, ...). 

\begin{table}[h]
    \caption{Comparing the impact of a single batch, offline fine-tuning, i.e. pre-training for the specific opening instead of from an empty board.
    }
    \label{tab:offline-fine-tuning}
    \vskip 0.1in
    \centering
    \begin{small}
        \begin{filecontents*}{tables/offline-fine-tuning.csv}
ID	w/o OL	w/ OL	T + w/o OL	T + w/ OL
JC	12,514	6,537	22,748	10,099
KE	18,771	3,336	2,248	2,417
DA	112,874	30,313	90,298	33,055
SE	64,227	12,465	28,905	42,522
\end{filecontents*}
\csvreader[
            tabular={crrrr},
            separator = tab,
            table head = \toprule & \multicolumn{2}{c}{w/o offline fine-tuning ($\theta_0$)} & \multicolumn{2}{c}{w/ offline fine-tuning ($\theta^{'}_0$)} \\
            \cmidrule(lr){2-3} \cmidrule(lr){4-5}
            & \textsc{baseline} & \textsc{online-cp} & \textsc{baseline} & \textsc{online-cp} \\ \midrule,
            late after last line = \\ \bottomrule,
        ]{tables/offline-fine-tuning.csv}{}{\csvcoli & \csvcolii & \csvcoliii & \csvcoliv & \csvcolv}
    \end{small}
\end{table}

Table \ref{tab:offline-fine-tuning} shows the times for solving these four openings with and without offline fine-tuning. 
The left two columns use $\theta_0$ while the right two columns use $\theta^{'}_0$.
With offline fine-tuning, the solving times for these openings generally decrease in the baseline solver, since the $\theta^{'}_0$ is specifically fine-tuned for each opening, but exceptions may still occur, as in opening \textit{JC}. 
However, when using $\theta^{'}_0$, the solving times for \textsc{online-cp} increase for opening \textit{JC}, \textit{DA}, and \textit{SE}.
This may be because $\theta^{'}_0$ only helps learn better heuristics for the opening positions, but does not always guarantee providing accurate heuristics for all varieties of positions during solving.
In addition, it is worth noting that although offline fine-tuned $\theta^{'}_0$ accelerates the solving time for the baseline solver, it is impractical since we cannot expect to pre-train $\theta^{'}_0$ for each opening, especially if our eventual goal is to solve complete games from an empty board outright.
In contrast, our online fine-tuning solver provides an automatic method that fine-tunes the PCN dynamically without too much extra computation cost.

%% file: figures/sgfs.tex
\begin{figure*}[h]
\centering
\subfloat[JA]{
\includegraphics[width=0.119\textwidth]{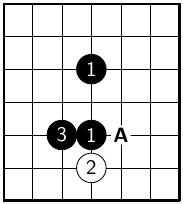}}
\subfloat[JB]{
\includegraphics[width=0.119\textwidth]{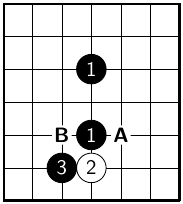}}
\subfloat[JC]{
\includegraphics[width=0.119\textwidth]{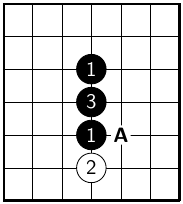}}
\subfloat[JD]{
\includegraphics[width=0.119\textwidth]{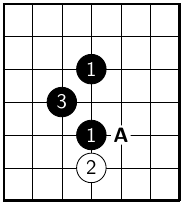}}
\subfloat[KA]{
\includegraphics[width=0.119\textwidth]{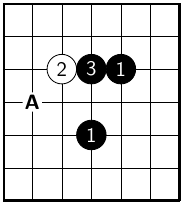}}
\subfloat[KB]{
\includegraphics[width=0.119\textwidth]{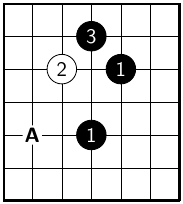}}
\subfloat[KC]{
\includegraphics[width=0.119\textwidth]{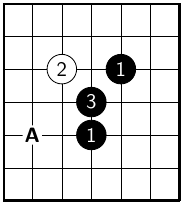}}
\subfloat[KD]{
\includegraphics[width=0.119\textwidth]{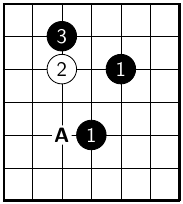}}
\\
\subfloat[KE]{
\includegraphics[width=0.119\textwidth]{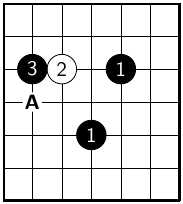}}
\subfloat[KF]{
\includegraphics[width=0.119\textwidth]{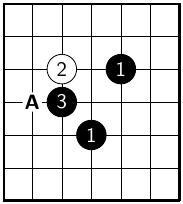}}
\subfloat[DA]{
\includegraphics[width=0.119\textwidth]{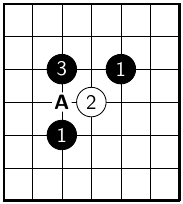}}
\subfloat[SA]{
\includegraphics[width=0.119\textwidth]{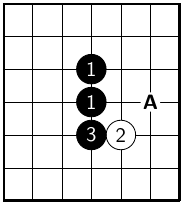}}
\subfloat[SB]{
\includegraphics[width=0.119\textwidth]{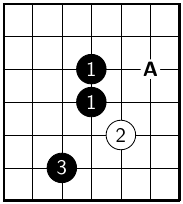}}
\subfloat[SC]{
\includegraphics[width=0.119\textwidth]{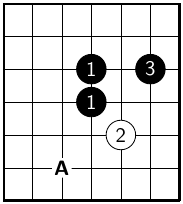}}
\subfloat[SD]{
\includegraphics[width=0.119\textwidth]{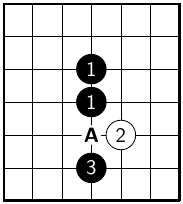}}
\subfloat[SE]{
\includegraphics[width=0.119\textwidth]{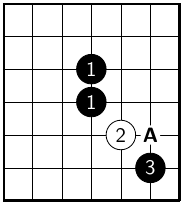}}

\caption{The solutions of the next winning move for 16 7x7 Killall-Go openings. For each opening, ``A'' and ``B'' represents the winning move found by the baseline solver and the online fine-tuning solver respectively. If both solvers solve the opening with the same winning move, only ``A'' is shown on the board.}
\label{fig:winning_moves}
\end{figure*}

%% file: figures/detailed-position-length.tex
\begin{figure}[!h]
    \captionsetup[subfigure]{justification=centering}
    \centering
    \subfloat[JA]{
        \begin{tikzpicture}[scale=0.5]
        \begin{axis} [
            xlabel = Iteration,
            ylabel = Length,
            width = \columnwidth,
            height = 0.5\columnwidth,
            compat = newest,
            grid style = dashed,
            ymajorgrids = true,
            legend pos = north west,
            enlarge x limits = 0.02,
            mark size = 1pt,
        ]
        \addplot+ [
            mark = x,
        ] table [
            x = Iteration,
            y = JA,
            col sep = tab
        ] {figures/position-length.csv};
        \addlegendentry{JA}
        \end{axis}
        \end{tikzpicture}
        \label{fig:position-length-detailed-JA}
    }
    \subfloat[JB]{
        \begin{tikzpicture}[scale=0.5]
        \begin{axis} [
            xlabel = Iteration,
            ylabel = Length,
            width = \columnwidth,
            height = 0.5\columnwidth,
            compat = newest,
            grid style = dashed,
            ymajorgrids = true,
            legend pos = north west,
            enlarge x limits = 0.02,
            mark size = 1pt,
        ]
        \addplot+ [
            mark = x,
        ] table [
            x = Iteration,
            y = JB,
            col sep = tab
        ] {figures/position-length.csv};
        \addlegendentry{JB}
        \end{axis}
        \end{tikzpicture}
        \label{fig:position-length-detailed-JB}
    }

    \subfloat[JC]{
        \begin{tikzpicture}[scale=0.5]
        \begin{axis} [
            xlabel = Iteration,
            ylabel = Length,
            width = \columnwidth,
            height = 0.5\columnwidth,
            compat = newest,
            grid style = dashed,
            ymajorgrids = true,
            legend pos = north west,
            enlarge x limits = 0.02,
            mark size = 1pt,
        ]
        \addplot+ [
            mark = x,
        ] table [
            x = Iteration,
            y = JC,
            col sep = tab
        ] {figures/position-length.csv};
        \addlegendentry{JC}
        \end{axis}
        \end{tikzpicture}
        \label{fig:position-length-detailed-JC}
    }
    \subfloat[JD]{
        \begin{tikzpicture}[scale=0.5]
        \begin{axis} [
            xlabel = Iteration,
            ylabel = Length,
            width = \columnwidth,
            height = 0.5\columnwidth,
            compat = newest,
            grid style = dashed,
            ymajorgrids = true,
            legend pos = north west,
            enlarge x limits = 0.02,
            mark size = 1pt,
        ]
        \addplot+ [
            mark = x,
        ] table [
            x = Iteration,
            y = JD,
            col sep = tab
        ] {figures/position-length.csv};
        \addlegendentry{JD}
        \end{axis}
        \end{tikzpicture}
        \label{fig:position-length-detailed-JD}
    }

    \subfloat[KA]{
        \begin{tikzpicture}[scale=0.5]
        \begin{axis} [
            xlabel = Iteration,
            ylabel = Length,
            width = \columnwidth,
            height = 0.5\columnwidth,
            compat = newest,
            grid style = dashed,
            ymajorgrids = true,
            legend pos = north west,
            enlarge x limits = 0.02,
            mark size = 1pt,
        ]
        \addplot+ [
            mark = x,
        ] table [
            x = Iteration,
            y = KA,
            col sep = tab
        ] {figures/position-length.csv};
        \addlegendentry{KA}
        \end{axis}
        \end{tikzpicture}
        \label{fig:position-length-detailed-KA}
    }
    \subfloat[KB]{
        \begin{tikzpicture}[scale=0.5]
        \begin{axis} [
            xlabel = Iteration,
            ylabel = Length,
            width = \columnwidth,
            height = 0.5\columnwidth,
            compat = newest,
            grid style = dashed,
            ymajorgrids = true,
            legend pos = north west,
            enlarge x limits = 0.02,
            mark size = 1pt,
        ]
        \addplot+ [
            mark = x,
        ] table [
            x = Iteration,
            y = KB,
            col sep = tab
        ] {figures/position-length.csv};
        \addlegendentry{KB}
        \end{axis}
        \end{tikzpicture}
        \label{fig:position-length-detailed-KB}
    }

    \subfloat[KC]{
        \begin{tikzpicture}[scale=0.5]
        \begin{axis} [
            xlabel = Iteration,
            ylabel = Length,
            width = \columnwidth,
            height = 0.5\columnwidth,
            compat = newest,
            grid style = dashed,
            ymajorgrids = true,
            legend pos = north west,
            enlarge x limits = 0.02,
            mark size = 1pt,
        ]
        \addplot+ [
            mark = x,
        ] table [
            x = Iteration,
            y = KC,
            col sep = tab
        ] {figures/position-length.csv};
        \addlegendentry{KC}
        \end{axis}
        \end{tikzpicture}
        \label{fig:position-length-detailed-KC}
    }
    \subfloat[KD]{
        \begin{tikzpicture}[scale=0.5]
        \begin{axis} [
            xlabel = Iteration,
            ylabel = Length,
            width = \columnwidth,
            height = 0.5\columnwidth,
            compat = newest,
            grid style = dashed,
            ymajorgrids = true,
            legend pos = north west,
            enlarge x limits = 0.02,
            mark size = 1pt,
        ]
        \addplot+ [
            mark = x,
        ] table [
            x = Iteration,
            y = KD,
            col sep = tab
        ] {figures/position-length.csv};
        \addlegendentry{KD}
        \end{axis}
        \end{tikzpicture}
        \label{fig:position-length-detailed-KD}
    }
    
    \subfloat[KE]{
        \begin{tikzpicture}[scale=0.5]
        \begin{axis} [
            xlabel = Iteration,
            ylabel = Length,
            width = \columnwidth,
            height = 0.5\columnwidth,
            compat = newest,
            grid style = dashed,
            ymajorgrids = true,
            legend pos = north west,
            enlarge x limits = 0.02,
            mark size = 1pt,
        ]
        \addplot+ [
            mark = x,
        ] table [
            x = Iteration,
            y = KE,
            col sep = tab
        ] {figures/position-length.csv};
        \addlegendentry{KE}
        \end{axis}
        \end{tikzpicture}
        \label{fig:position-length-detailed-KE}
    }
    \subfloat[KF]{
        \begin{tikzpicture}[scale=0.5]
        \begin{axis} [
            xlabel = Iteration,
            ylabel = Length,
            width = \columnwidth,
            height = 0.5\columnwidth,
            compat = newest,
            grid style = dashed,
            ymajorgrids = true,
            legend pos = north west,
            enlarge x limits = 0.02,
            mark size = 1pt,
        ]
        \addplot+ [
            mark = x,
        ] table [
            x = Iteration,
            y = KF,
            col sep = tab
        ] {figures/position-length.csv};
        \addlegendentry{KF}
        \end{axis}
        \end{tikzpicture}
        \label{fig:position-length-detailed-KF}
    }
    \caption{Average length of critical positions for each opening.}
    \label{fig:position-length-detailed}
\end{figure}

\begin{figure}[h]
    \ContinuedFloat
    \captionsetup[subfigure]{justification=centering}
    \centering
    \subfloat[DA]{
        \begin{tikzpicture}[scale=0.5]
        \begin{axis} [
            xlabel = Iteration,
            ylabel = Length,
            width = \columnwidth,
            height = 0.5\columnwidth,
            compat = newest,
            grid style = dashed,
            ymajorgrids = true,
            legend pos = north west,
            enlarge x limits = 0.02,
            mark size = 1pt,
        ]
        \addplot+ [
            mark = x,
        ] table [
            x = Iteration,
            y = DA,
            col sep = tab
        ] {figures/position-length.csv};
        \addlegendentry{DA}
        \end{axis}
        \end{tikzpicture}
        \label{fig:position-length-detailed-DA}
    }
    \subfloat[SA]{
        \begin{tikzpicture}[scale=0.5]
        \begin{axis} [
            xlabel = Iteration,
            ylabel = Length,
            width = \columnwidth,
            height = 0.5\columnwidth,
            compat = newest,
            grid style = dashed,
            ymajorgrids = true,
            legend pos = north west,
            enlarge x limits = 0.02,
            mark size = 1pt,
        ]
        \addplot+ [
            mark = x,
        ] table [
            x = Iteration,
            y = SA,
            col sep = tab
        ] {figures/position-length.csv};
        \addlegendentry{SA}
        \end{axis}
        \end{tikzpicture}
        \label{fig:position-length-detailed-SA}
    }
    
    \subfloat[SB]{
        \begin{tikzpicture}[scale=0.5]
        \begin{axis} [
            xlabel = Iteration,
            ylabel = Length,
            width = \columnwidth,
            height = 0.5\columnwidth,
            compat = newest,
            grid style = dashed,
            ymajorgrids = true,
            legend pos = north west,
            enlarge x limits = 0.02,
            mark size = 1pt,
        ]
        \addplot+ [
            mark = x,
        ] table [
            x = Iteration,
            y = SB,
            col sep = tab
        ] {figures/position-length.csv};
        \addlegendentry{SB}
        \end{axis}
        \end{tikzpicture}
        \label{fig:position-length-detailed-SB}
    }
    \subfloat[SC]{
        \begin{tikzpicture}[scale=0.5]
        \begin{axis} [
            xlabel = Iteration,
            ylabel = Length,
            width = \columnwidth,
            height = 0.5\columnwidth,
            compat = newest,
            grid style = dashed,
            ymajorgrids = true,
            legend pos = north west,
            enlarge x limits = 0.02,
            mark size = 1pt,
        ]
        \addplot+ [
            mark = x,
        ] table [
            x = Iteration,
            y = SC,
            col sep = tab
        ] {figures/position-length.csv};
        \addlegendentry{SC}
        \end{axis}
        \end{tikzpicture}
        \label{fig:position-length-detailed-SC}
    }
    
    \subfloat[SD]{
        \begin{tikzpicture}[scale=0.5]
        \begin{axis} [
            xlabel = Iteration,
            ylabel = Length,
            width = \columnwidth,
            height = 0.5\columnwidth,
            compat = newest,
            grid style = dashed,
            ymajorgrids = true,
            legend pos = north west,
            enlarge x limits = 0.02,
            mark size = 1pt,
        ]
        \addplot+ [
            mark = x,
        ] table [
            x = Iteration,
            y = SD,
            col sep = tab
        ] {figures/position-length.csv};
        \addlegendentry{SD}
        \end{axis}
        \end{tikzpicture}
        \label{fig:position-length-detailed-SD}
    }
    \subfloat[SE]{
        \begin{tikzpicture}[scale=0.5]
        \begin{axis} [
            xlabel = Iteration,
            ylabel = Length,
            width = \columnwidth,
            height = 0.5\columnwidth,
            compat = newest,
            grid style = dashed,
            ymajorgrids = true,
            legend pos = north west,
            enlarge x limits = 0.02,
            mark size = 1pt,
        ]
        \addplot+ [
            mark = x,
        ] table [
            x = Iteration,
            y = SE,
            col sep = tab
        ] {figures/position-length.csv};
        \addlegendentry{SE}
        \end{axis}
        \end{tikzpicture}
        \label{fig:position-length-detailed-SE}
    }
    \caption{Average length of critical positions for each opening.}
\end{figure}